\documentclass[runningheads]{llncs}
\usepackage{graphicx}
\usepackage{xcolor}
\usepackage{algorithm}
\usepackage{amsmath}
\usepackage{amssymb}
\usepackage{siunitx}
\usepackage{booktabs}
\usepackage{mathtools}
\usepackage{xcolor}
\usepackage[noend]{algpseudocode}

\begin{document}
\title{Inceptive Event Time-Surfaces for Object Classification using Neuromorphic Cameras\thanks{This work is funded in part by Ford University Research Program.}}
\titlerunning{Inceptive Event Time-Surfaces}
%
\author{R. Wes Baldwin \and
Mohammed Almatrafi \and
Jason R. Kaufman \and
Vijayan Asari \and
Keigo Hirakawa}
\authorrunning{R. Baldwin et al.}
%
\institute{University of Dayton, Dayton OH 45469, USA \email{\{baldwinr2,almatrafim2,jkaufman1,vasari1,khirakawa1\}@udayton.edu}\\
\url{https://www.udayton.edu/engineering/research/centers/vision\_lab}\\
\url{http://issl.udayton.edu}\\}
\maketitle              
\begin{abstract}
This paper presents a novel fusion of low-level approaches for dimensionality reduction into an effective approach for high-level objects in neuromorphic camera data called Inceptive Event Time-Surfaces (IETS). IETSs overcome several limitations of conventional time-surfaces by increasing robustness to noise, promoting spatial consistency, and improving the temporal localization of (moving) edges. Combining IETS with transfer learning improves state-of-the-art performance on the challenging problem of object classification utilizing event camera data.

\keywords{Object Classification \and Dynamic Vision \and Neuromorphic Vision \and Dimensionality Reduction.}
\end{abstract}
\section{Introduction}

A standard image sensor is comprised of an array of Active Pixel Sensors (APS). Each APS circuit reports the pixel intensity of the image formed at the focal plane by cycling between a period of integration (wherein photons are collected and counted by each pixel detector) and a readout period (where digital counts are combined from all pixels to form a single frame). Motion detected and estimated across frames has useful applications in computer vision tasks. Unfortunately, detecting fast moving objects can be challenging due to the limitations of the integration and read out circuit. Object motion that is too fast relative to the integration period induces blurring and other artifacts. Additionally, since all pixels have a single exposure setting, parts of the scene may be underexposed while other parts are saturated. Both of these issues degrade the image quality of the captured video frames, reducing our ability to detect or recognize objects by their shapes or their motions. While high-speed cameras with very fast frame rates can resolve blur issues, they are expensive, consume lots of power, generate large amounts of data, and require adjusting exposure settings.

Event-based cameras were engineered to overcome these limitations of the APS circuitry found on conventional framing cameras. As described below, these neuromorphically inspired cameras can operate at extremely high temporal resolution ($>$800kHz), low latency (20 microseconds), wide dynamic range ($>120$dB), and low power (30mW). They report only changes in the pixel intensity, requiring a new set of techniques to perform basic image processing and computer vision tasks---examples include optical flow \cite{bardow2016simultaneous,haessig2018spiking}, feature extraction \cite{barranco2015contour,mitrokhin2018event,mueggler2017event}, gesture recognition \cite{amir2017low,lee2014real}, and object recognition \cite{barua2016direct,orchard2015hfirst}. 

``Time-surface'' is one such technique with proven usefulness in pattern recognition by encoding the event-time as an intensity \cite{lagorce2017hots}. However, time-surfaces are sensitive to noise and to multiple events corresponding to the same image edge with some latency when the intensity changes are large. Both have an effect on time-surfaces similar to the ways that blurring affects APS data. An improved time-surface technique called Filtered Surface of Active Events (FSAE) \cite{alzugaray2018asynchronous} was introduced in a corner detection and tracking algorithm. FSAE yields an improved time-surface by only utilizing the initial event of a series---effectively removing events corresponding to the same edge. Yet, while FSAE is shown to be very effective for representing simple features such as corners, object classification tasks deal with significantly more complex objects.

In this work, we propose IETS, aimed at extracting noise-robust, low-latency features that correspond to complex object edge contours over a temporal window. IETS extends FSAE to achieve higher object recognition accuracy while removing over 70\% of FSAE events. We verify the effectiveness of our object classification framework on multiple datasets.

\subsection{Event Cameras}

\begin{figure}[t]
    \begin{minipage}[b]{0.48\linewidth}
        \centering
        \includegraphics[width=1\linewidth]{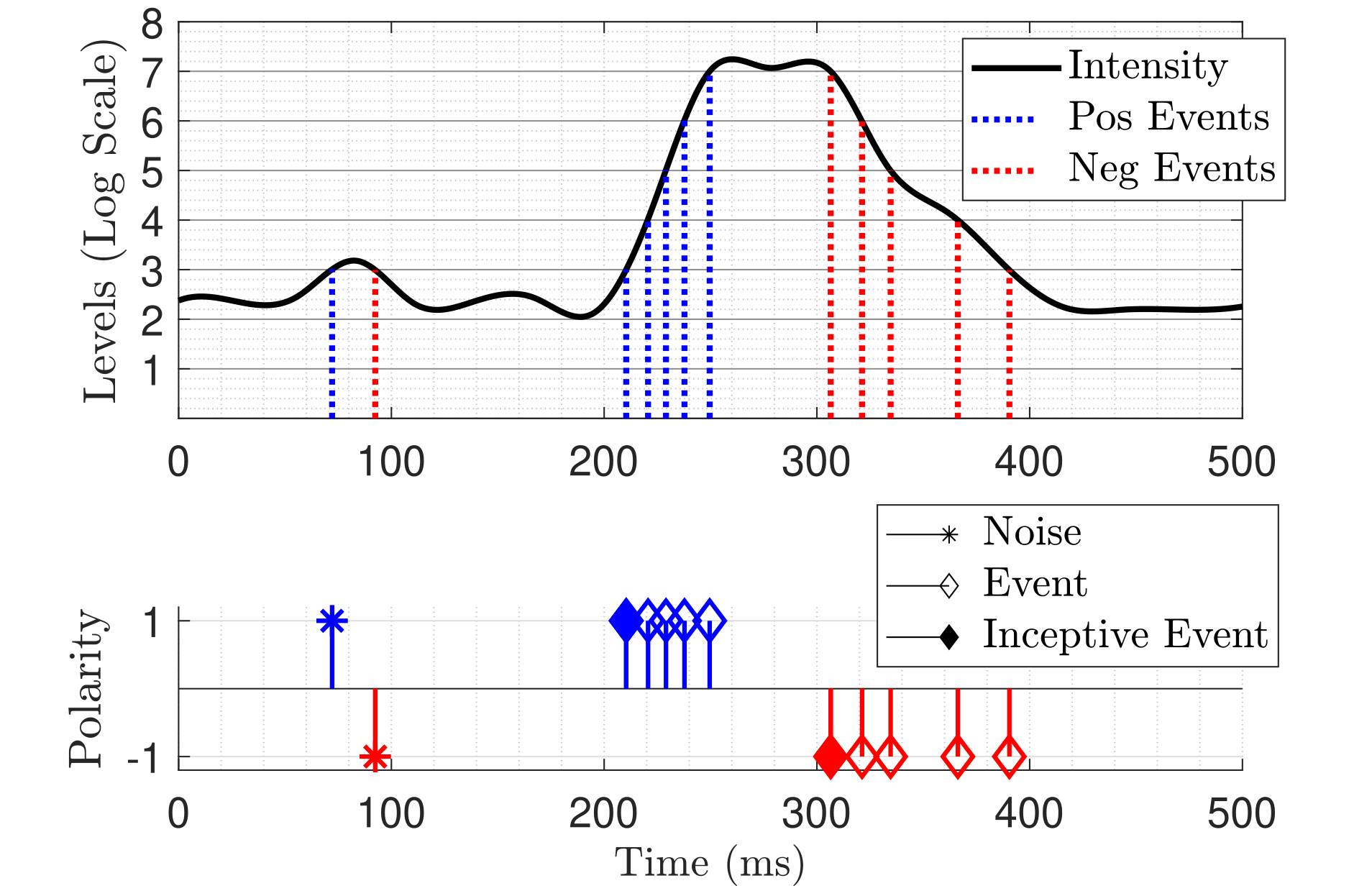}\\(a)
    \end{minipage}
    \hfill
    \begin{minipage}[b]{0.48\linewidth}
        \centering
        \includegraphics[width=1\linewidth]{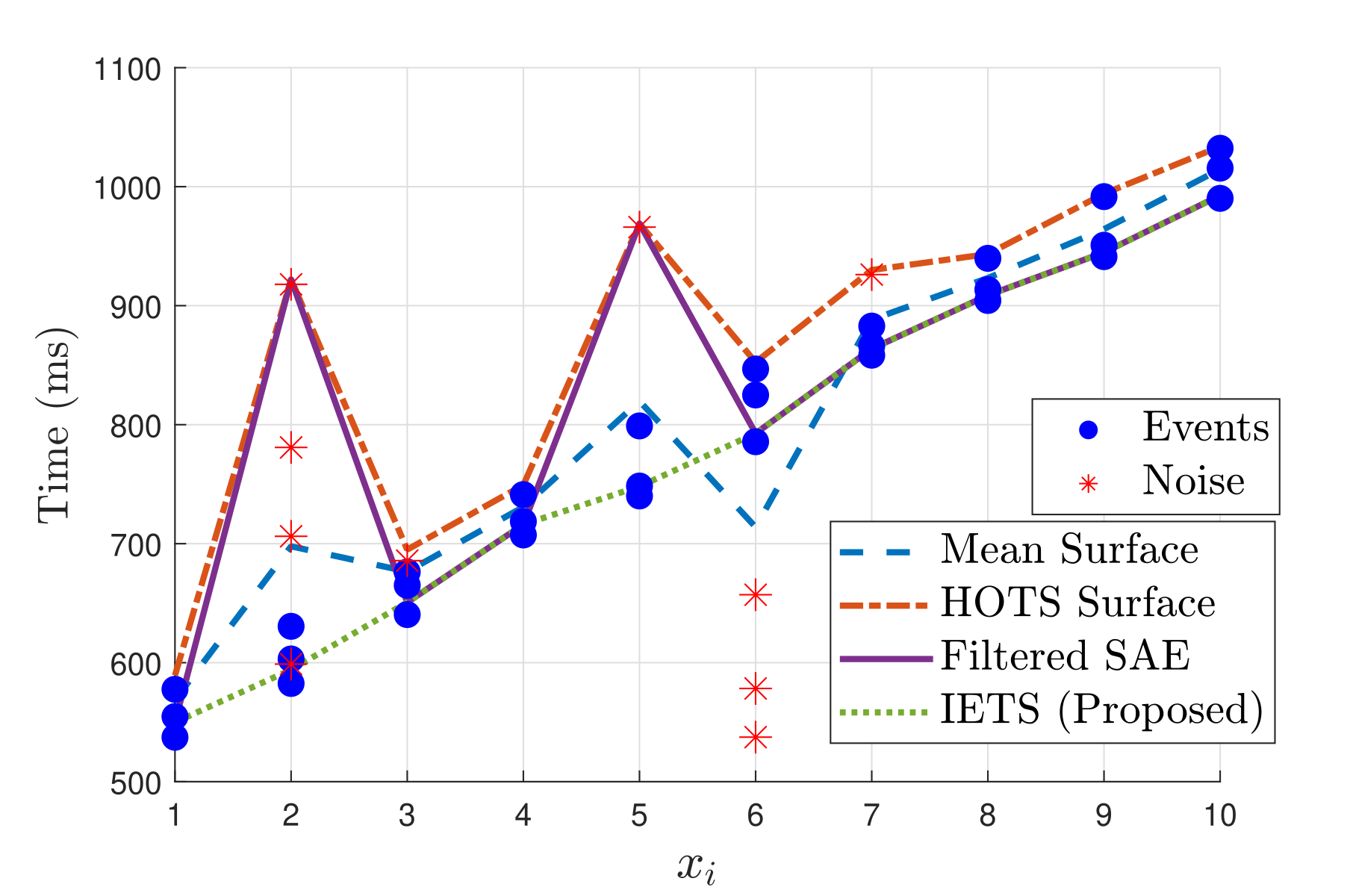}\\(b)
    \end{minipage}
    \caption{Event Generation. \textbf{(a)} On a per pixel level, intensity variations trigger events at each log-scaled level crossing. The first event in a series of consecutive events is called an Inceptive Event. \textbf{(b)} Time-Surface generation in the presence of noise.}
    \label{fig:eventplots}
\end{figure}

Each event-based camera pixel operates asynchronously with no notion of frame rate across the focal plane. Instead of a fixed integration time, pixels generate events only when the rate of detected photons varies above or below a predefined threshold. A log-based threshold gives the event camera an extreme dynamic range. If the scene is changing slowly, the sensor naturally compresses the data since few events are generated. In contrast, fast moving objects trigger events almost instantaneously---allowing object tracking within microseconds. Example event generation for a single pixel is illustrated in Fig.~\ref{fig:eventplots}(a). 

In a Prophesee Asynchronous Time-based Image Sensor, used in N-CARS \cite{sironi2018hats}, each event comprises a row, column, time, and polarity. Row and column are the pixel coordinates. The time entry records when the change was detected in microseconds, and the polarity is a binary value indicating if the intensity increased or decreased. 

Event camera data is often noisy and requires filtering for many applications. Previous algorithms rely on the assumption that when a pixel is triggered, neighboring pixels are also activated \cite{czech2016evaluating,padala2018noise} and large intensity changes generate multiple events at a single pixel. These assumptions motivate the use of spatial-temporal density as a way to isolate valid events from noise, but this approach fails when motion is slow (i.e. sparse valid events are removed as noise) and when noise is high (i.e. dense noise mislabeled as real events).

\subsection{N-CARS Dataset}

The N-CARS dataset is a large, real-world, event-based, public dataset for car classification. It is composed of 12,336 car samples and 11,693 non-cars samples (background). The camera was mounted behind the windshield of a car and gives a view similar to what the driver would see. Each sample contains exactly 100 milliseconds of data with 500 to 59,249 events per sample.

Fig.~\ref{fig:car_example} shows a sequence from N-CARS; each point in the three dimensional cube (2D space, 1D time) represents a reported event. Object velocity can be inferred when this cube is viewed from the time-space plane (Fig.~\ref{fig:car_example}(a)), while the object shape is better identifiable from the 2D space plane (Fig.~\ref{fig:car_example}(b)). Spiral patterns near the rear wheel of the car highlight high-speed rotational motion---a challenging set of relevant features to preserve during dimensionality reduction.

\section{Related Works}

\begin{figure*}[t]
    \begin{minipage}[b]{0.48\linewidth}
        \centering
        \includegraphics[width=1\linewidth]{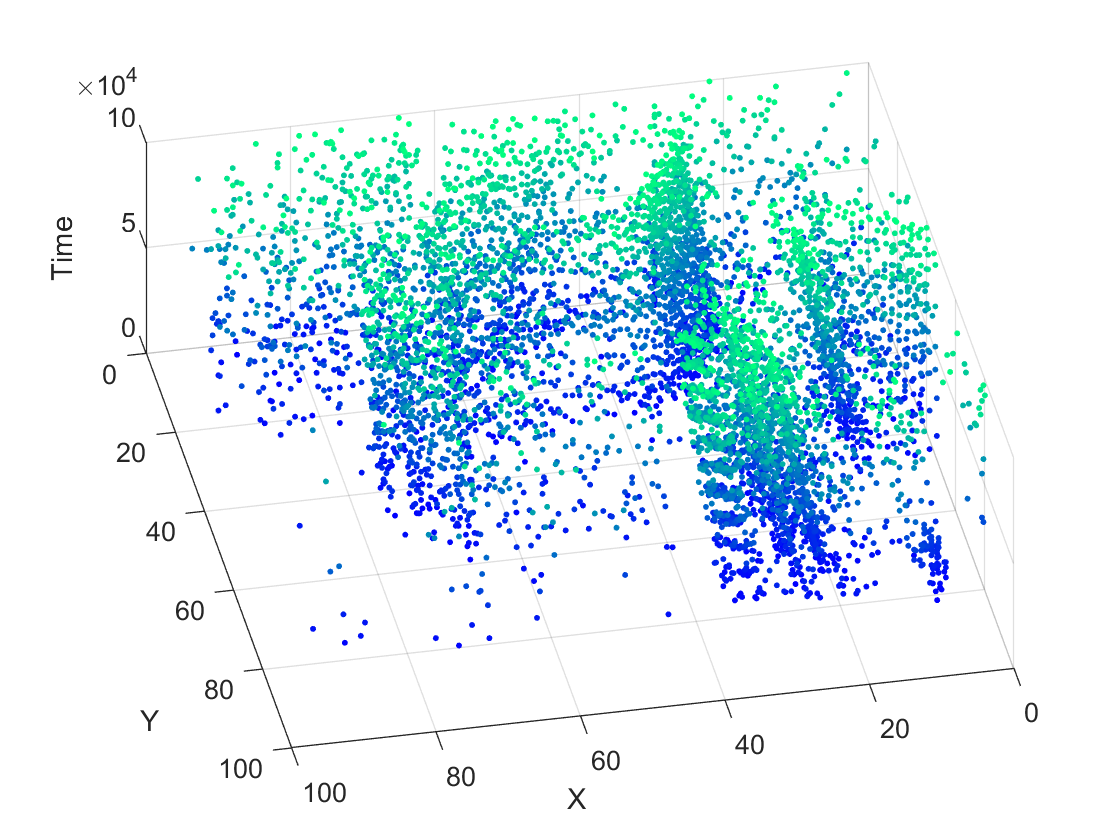}\\(a)
    \end{minipage}
    \hfill
    \begin{minipage}[b]{0.48\linewidth}
        \centering
        \includegraphics[width=1\linewidth]{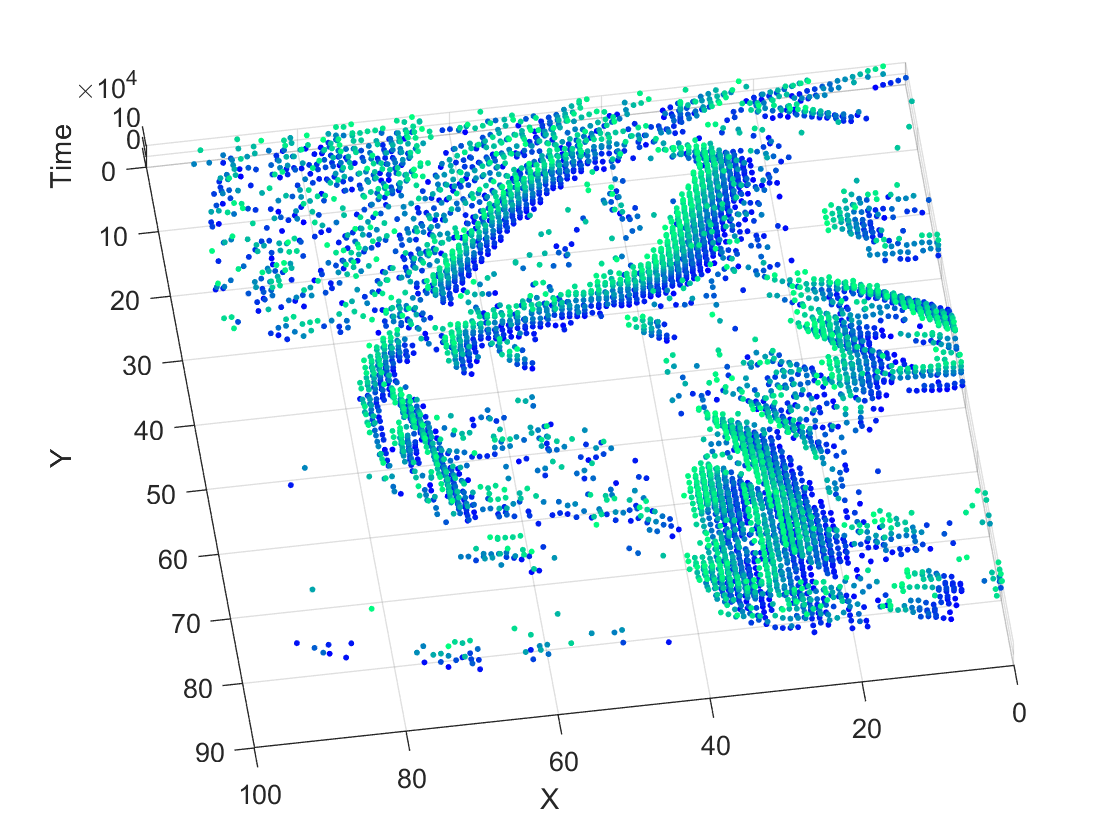}\\(b)
    \end{minipage}
    \caption{N-CARS Dataset Example. \textbf{(a)} 3D plot of event data colored by time. (Blue/old to green/new). \textbf{(b)} Same data viewed under different orientation.}
    \label{fig:car_example}
\end{figure*}

Object classification from event data is an active area of research. There are a number of applications that require feature extraction from the raw event detection camera data in order to carry out classification tasks. Time-surface is a technique used as an intermediary step to feature extraction by reducing the spatial-temporal structure in Fig.~\ref{fig:car_example} to a two dimensional image representation. More specifically, let $E$ denote a set comprised of events generated by an event detection camera sensor of frame size $M \times N$:
\begin{equation}
  \label{eventeqn}
  E(x,y) = \{(t_i,p_i)\}_{i=1}^I,
\end{equation}
where $x\in [1,...,N]$ and $y\in [1,...,M]$ represent the pixel coordinates in the frame; $p_i\in \{-1,1\}$ is the event polarity; and $t_i$ is the time of the event in microseconds. Additionally, let $T$ be an ordered set of event times for a single pixel $(x,y)$ with polarity $p$ be defined as:
\begin{equation}
  \label{timeeqn}
    T(x,y,p) = \{e_i \in E\;|\;p_i=p\}.
\end{equation}
Then the time-surface for each pixel $(x,y)$ with polarity $p$ is defined as \cite{lagorce2017hots}:
\begin{equation}
  \label{TSeqn}
  \mathcal{TS}\{T\}(x,y,p) = \operatorname{mean}\{T(x,y,p)\}=\frac{1}{\mid T\mid}\sum\limits_{(t_i,p_i)\in T(x,y,p)} {t_i}.
\end{equation}
Variations to time-surface can be implemented by replacing the ``mean'' operator in \eqref{TSeqn} with minimum, maximum, median, etc.

Time-surface has been used successfully in object recognition tasks. For example, Hierarchy of Time-Surfaces (HOTS) \cite{lagorce2017hots} utilized straightforward time-surfaces for feature generation, but it did not attempt to limit the impact of noise directly, instead relying on clustering. While this method performed well on simple shapes like numbers and letters, it does not extend well to more complex-shaped objects with wider variations (like cars). 

The Histograms of Averaged Time-Surfaces (HATS) algorithm \cite{sironi2018hats} localizes the motion vector representation for a specific region of the sensor (cell) using a region-based time-surface. This improved robustness to noise by averaging across the reported times of the events within each cell. A major disadvantage to HATS is the loss of fine spatial features, which is exacerbated by the low sensor resolution of current event cameras.

FSAE is a method to directly improve time-surface by eliminating redundant events\cite{alzugaray2018asynchronous}. The FSAE filter is defined as:
\begin{equation}
  \label{FASEeqn}
    \mathcal{FSAE}(x,y,p) = \{t_i \in T(x,y,p)\;|\;(t_i-t_{i-1}) > \tau^-\},
\end{equation}
where $\tau^-$ is a pre-defined threshold. Intuitively, events occurring in succession typically correspond to the same edge, and so redundant events can be eliminated by discarding events that are not temporally separated from prior events.


\section{Proposed Method: Inceptive Event Time-Surfaces}

To advance object classification using event data, we propose a novel concept called Inceptive Event Time-Surfaces (IETS). IETS is an extension of FSAE aimed at improving dimensionality reduction and noise robustness. IETS retains features critical to object classification (i.e. corners and edges) by fitting time-surfaces to a subset of events. Unlike previous approaches that focused on generating handcrafted features from noisy event data, IETS uses deep convolutional neural networks (CNNs) to learn features from time-surface images with less noise. As demonstrated by the experiments using the N-CARS, IETS combined with CNNs achieves a new state-of-the-art in classification performance.

We begin by the observation that a single log-intensity change often trigger multiple events in temporal sequence. As shown by Fig.~\ref{fig:eventplots}(a), the first event indicates an ``arrival'' of an edge, which we refer to as an ``inceptive event" (IE). Intuitively, IEs describe the shape of the moving object within the scene. On the other hand, the subsequent events correspond to the magnitude of the log-intensity change, which we refer to as ``scaling events.'' As such, edge magnitude as indicated by successive scaling events do not necessarily describe the edge shape well. The comparison between inceptive and scaling events in Fig.~\ref{fig:eventplots}(a) make this clear. While scaling events are more useful for intensity-based inferences, the effect the latency (relative to the edge arrival) has on the time-surface is similar to image blur. Furthermore, scaling events are subject to degradation by two hardware designs: a low-pass filter and a regulated ``refractory period''---a period of time after an event trigger that a pixel must wait before triggering again (due to the limitations of read out and reset circuits).

Object detection tasks require a clear representation of the object boundaries that define the shape of the object-of-interest. Recall \eqref{timeeqn}. To successfully filter events prior to time-surface generation, we propose the following:
\begin{equation}
  \label{IEeqn}
  \mathcal{IE}(x,y,p) = \{t_i \in T(x,y,p) | (t_i-t_{i-1}) > \tau^-\;\land\;(t_{i+1}-t_i) < \tau^+\},
\end{equation}
where $\tau^+$ and $\tau^-$ are predefined threshold parameters. One may notice that by comparing \eqref{IEeqn} to \eqref{FASEeqn} that $\mathcal{IE}\subset\mathcal{FSAE}$, meaning there are necessarily fewer IEs than FSAE events. The proposed $\mathcal{IETS}$ is then defined as a time surface constructed from $\mathcal{IE}$:
\begin{equation}
  \label{IETSeqn}
  \mathcal{IETS}(x,y,p) = TS\{\mathcal{IE}\}(x,y,p).
\end{equation}

We propose to carry out the object classification by training a CNN on IETS surfaces. There are three input image channels to the proposed CNN. First two input channels are IETS surfaces of both polarities: $\mathcal{IETS}(x,y,+1)$ and $\mathcal{IETS}(x,y,-1)$, which are mapped to images of 8-bit intensity values. The third input channel is generated based on a simple count of unfiltered events (i.e.~$E(x,y)$) at each pixel. This channel can improve machine learning by acting as a weight for the other channels. All channels are scaled from 0 to 1, and pixels with no events in the entire dataset are set to zero. With $\tau^-=\tau^+=12ms$, IETS removes over 85\% of events in N-CARS. Discriminating noise from real events can be challenging, degrading time-surfaces significantly. Fig.~\ref{fig:eventplots}(b) highlights the effectiveness of IETS in removing noise while accurately fitting the time-surface, compared to other methods.

\begin{figure*}[t]
    \begin{minipage}[b]{0.32\linewidth}
        \centering
        \includegraphics[width=1\linewidth]{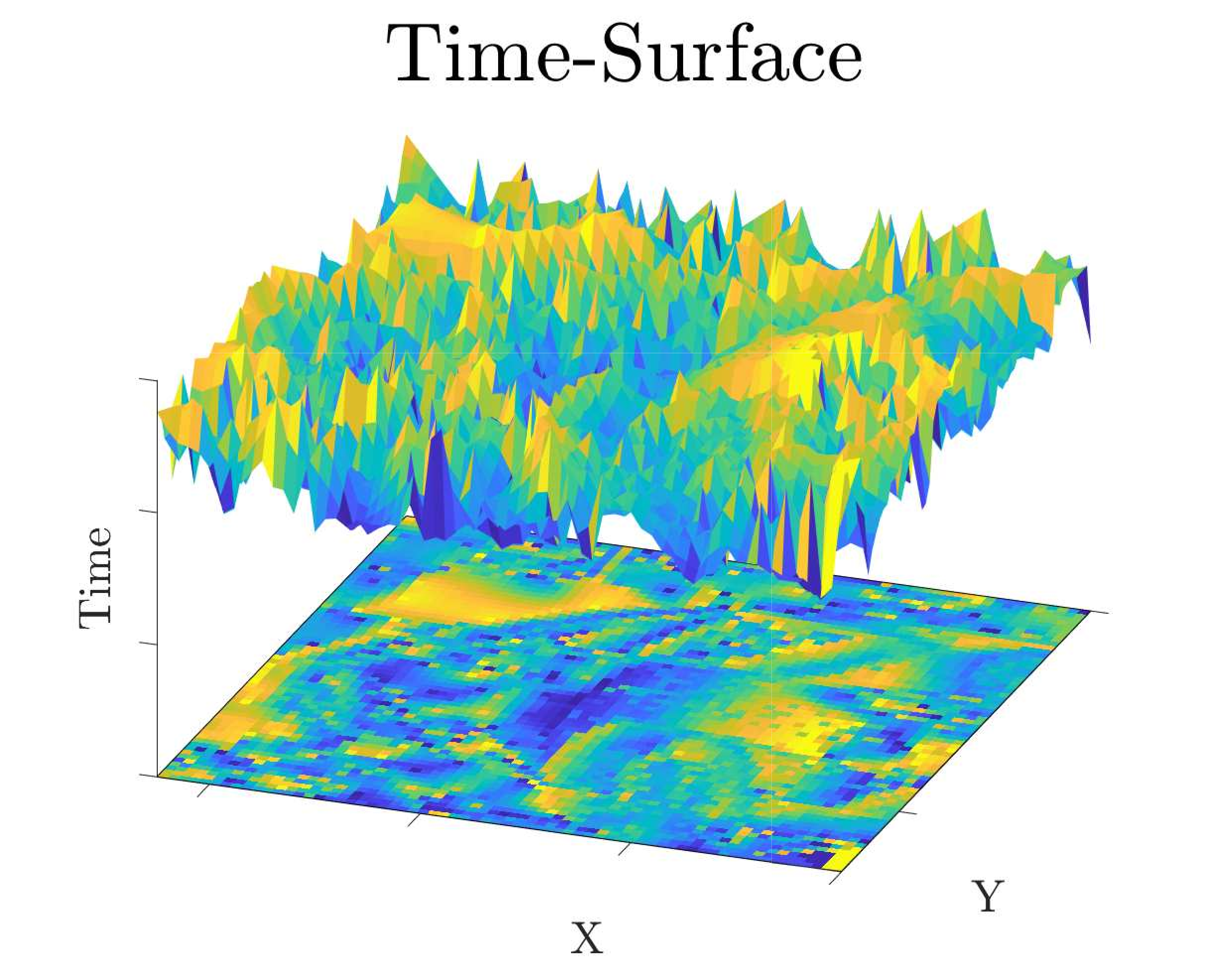}\\(a)
    \end{minipage}
    \hfill
    \begin{minipage}[b]{0.32\linewidth}
        \centering
        \includegraphics[width=1\linewidth]{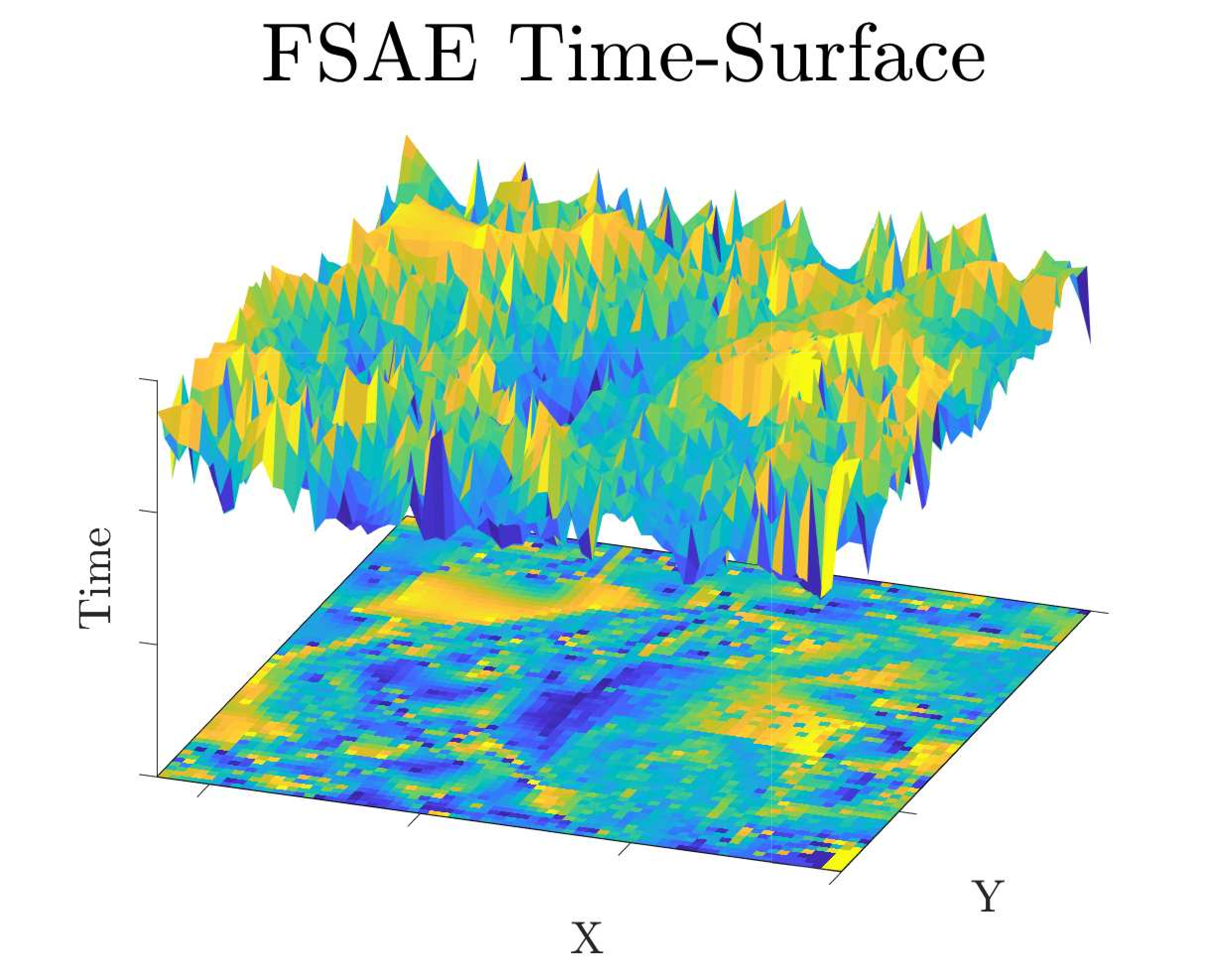}\\(b)
    \end{minipage}
    \hfill
    \begin{minipage}[b]{0.32\linewidth}
        \centering
        \includegraphics[width=1\linewidth]{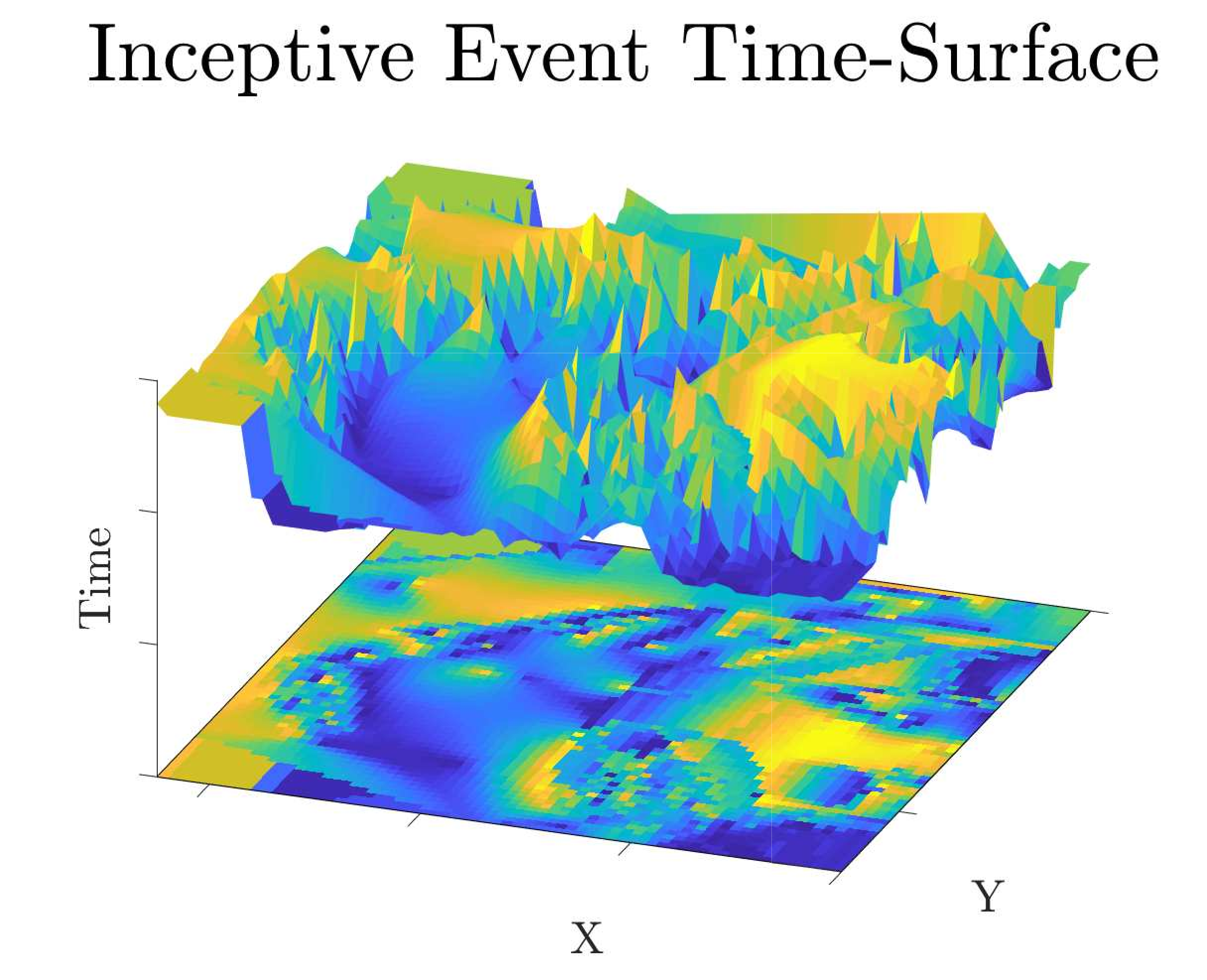}\\(c)
    \end{minipage}
    \caption{Time-Surface Visualization. \textbf{(a)} Noisy 2D time-surface (bottom) compiled from $\sim17k$ events represented as a 3D mesh (above) \textbf{(b)} Same visualization constructed from subset of $\sim8k$ FSAE events. \textbf{(c)} Same visualization constructed from subset of $\sim3k$ IETS events. IETS shows significantly less noise in time-surface, representing meaningful image features better than the unfiltered sensor events or FSAE events.}
    \label{fig:3dcar_example}
\end{figure*}

Due to the extremely sparse number of events ($<1k$) in some N-CARS datasets, likely captured during periods of little camera or target motion, IETS filtering occasionally makes object identification even more challenging. For that reason, if a pixel does not contain an IE, the mean time of all events for that pixel is used in its place. Although this reintroduces noise to each image, the overall classification accuracy on N-CARS improved by over 12\% when mean event time for non-IE data was appended. Additional data, even if very noisy, is preferred when using deep neural networks. Fig. \ref{fig:3dcar_example} highlights how effectively IETS can reduce dimensionality while at the same time removing noisy events.

Previous event-based features \cite{clady2017motion,lagorce2017hots} are limited in the same way as many custom-designed descriptors. Leveraging CNNs to learn optimal features is typically a superior approach over custom-designed features. Of course, deep convolutional neural networks currently require millions of labeled images---something that does not yet exist for event cameras. Since no vast archive of labeled event camera data exist, IETS images are generated in a way that makes them optimal to utilize transfer feature learning from millions of real-world images via GoogLeNet \cite{krizhevsky2012imagenet,szegedy2015going}. IETS is highly parallelizable and quick to train since transfer learning converges rapidly. IETS generates images at the full resolution of the event camera. This means resolution, which is typically poor for event cameras, is not lost prior to classification as with algorithms employing cells.

IETS has excellent performance as all events in a given time window are processed simultaneously---removing the requirement to iterate over each event. Additionally, a non-optimal implementation of IETS processed over 100k events/sec, significantly faster than real-time requirements.

\section{Experiments}

\begin{figure*}[t]
  \centering
  \begin{tabular}{c c c}
    \includegraphics[width=.40\linewidth]{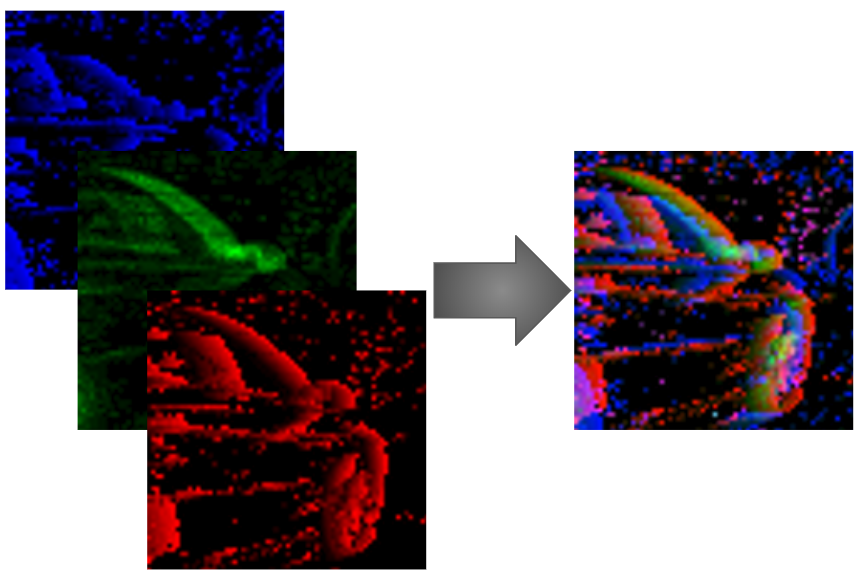} &
    \includegraphics[width=.27\linewidth]{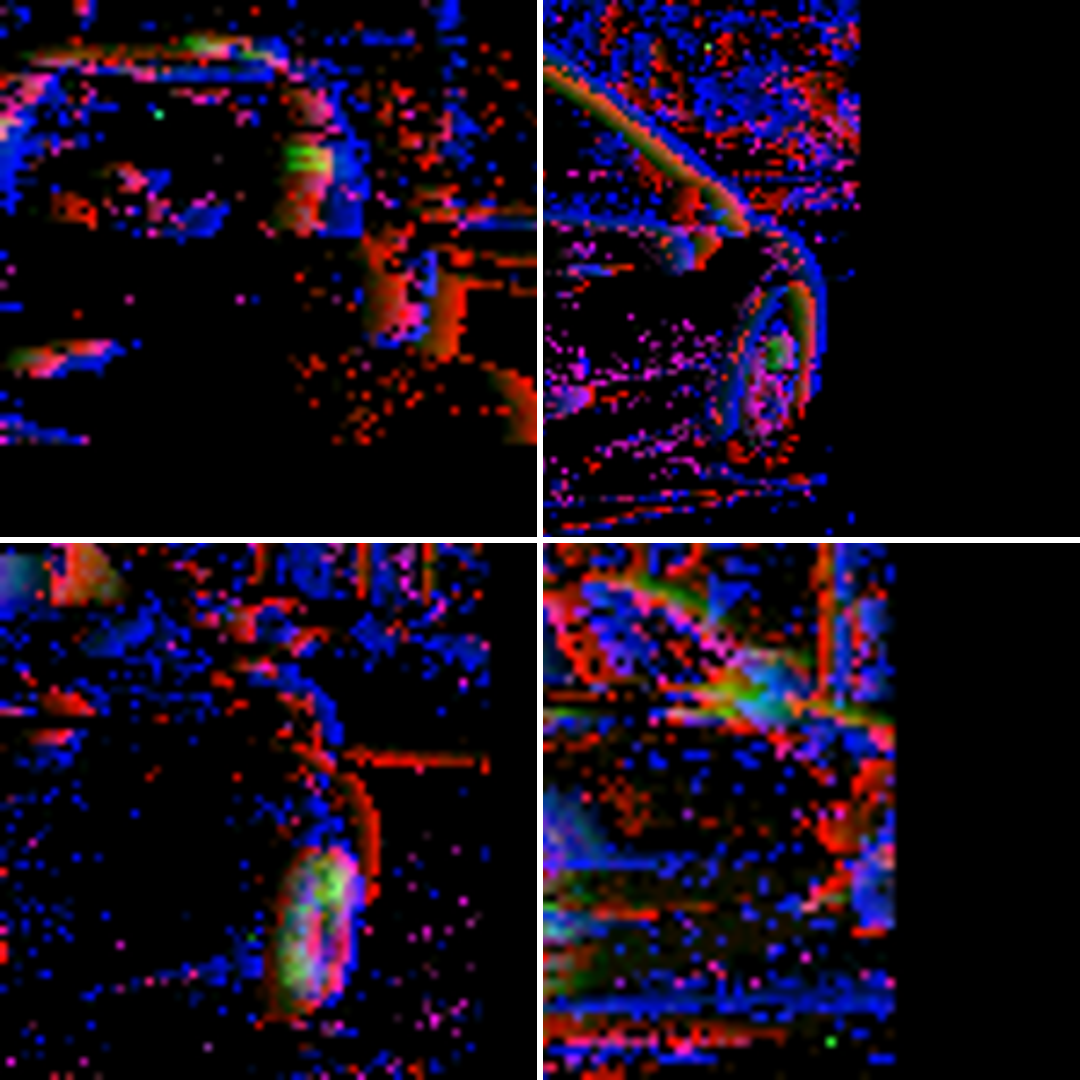} & 
    \includegraphics[width=.27\linewidth]{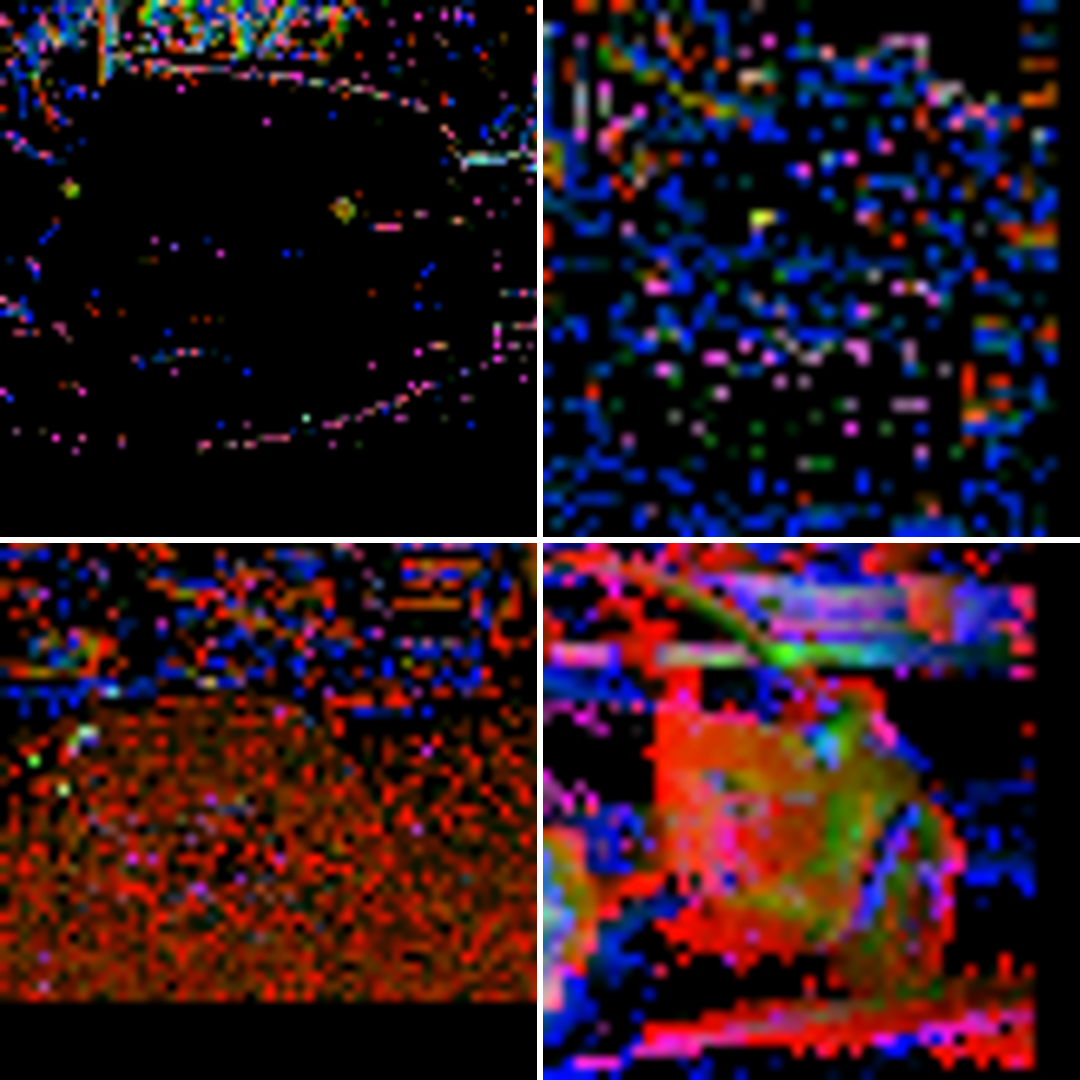} \\
    \small (a) & \small (b) & \small (c) \\
  \end{tabular}
  \caption{\textbf{(a)} Example input to CNN is two IETSs (positive/negative polarity) and the event count per pixel (shown here as RGB). Examples from N-CARS dataset that were \textbf{(b)} correctly and \textbf{(c)} incorrectly labeled as `cars.'}
  \label{fig:montage}
\end{figure*}

Each N-CARS sample was processed into an image using IETS. Examples from IETS processing are shown in Fig. \ref{fig:montage}. Algorithm evaluation was accomplished via the standard metrics of accuracy rate and Area Under Curve (AUC).

The maximum score was produced after augmenting the training data by using IETS images that had also been flipped. The maximum accuracy score obtained by IETS was $0.973$. Comparison to other state-of-the-art algorithms is shown in Table \ref{tab:accuracy}, and is a considerable improvement over the HATS published score of $0.902$. AUC also improved from $0.945$ to $0.997$. To ensure performance gains were not entirely from replacing the Support Vector Machine (SVM) with a CNN, HATS features were used to train the same GoogLeNet architecture. These results are also included Table \ref{tab:accuracy} as HATS/CNN. Additionally, to show the improvement IETS offers in generating a time-surface, FSAE images were used to train the architecture and are also included for comparison.

\begin{table}[htbp]
  \centering
  \caption{Classification results on N-CARS.}
    \setlength{\tabcolsep}{4pt}
    \begin{tabular}{lccccccc}
    \toprule
    \textbf{Algorithm} & \textbf{H-First} & \textbf{HOTS} & \textbf{Gabor} & \textbf{HATS} & \textbf{HATS} & \textbf{FSAE} & \textbf{IETS} \\ 
    \midrule
    Classifier & SNN & SVM & SNN & SVM & CNN & CNN & \textbf{CNN} \\
    Accuracy & 0.561 & 0.624 & 0.789 & 0.902 & 0.929 & 0.961 & \textbf{0.976} \\
    AUC & 0.408 & 0.568 & 0.735 & 0.945 & 0.984 & 0.993 & \textbf{0.997} \\
    \bottomrule
    \end{tabular}%
  \label{tab:accuracy}%
\end{table}%

\begin{figure}[t]
    \begin{minipage}[b]{0.32\linewidth}
        \centering
        \includegraphics[width=1\linewidth]{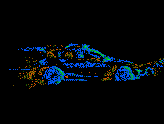}
    \end{minipage}
    \hfill
    \begin{minipage}[b]{0.32\linewidth}
        \centering
        \includegraphics[width=1\linewidth]{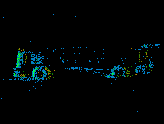}
    \end{minipage}
    \hfill
    \begin{minipage}[b]{0.32\linewidth}
        \centering
        \includegraphics[width=1\linewidth]{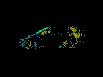}
    \end{minipage}
    \caption{Three IETS images generated from data collected near the University of Dayton used for additional testing. Data included multiple cars, buses, and trucks.}
    \label{fig:sideCars}
\end{figure}

To further test the results from IETS, an IniVation Davis Dynamic Vision Sensor (DVS) 240C was used to collect cars driving near the University of Dayton. This dataset was significantly different in the fact targets were acquired using a camera from a different manufacturer, at a further range, images were uncropped, and the camera was stationary. The vehicles collected were side on as shown in Fig. \ref{fig:sideCars}. Seven datasets were recorded with durations ranging from 2.76 to 8.30 seconds---resulting in 5,236 samples. Using four datasets for training and three for testing resulted in a classification accuracy of 0.9951 and AUC score of 0.9999. Although the dataset proved less challenging, the results indicate that supplementing with additional variation in sensor models, viewing angles, and camera positions will allow the algorithm to extend to more general use cases.

\section{Conclusion and Future Work}

Overall, there are a wide range of future applications for event-based sensors due to their speed, size, low memory requirements, and high dynamic range. This paper presents an algorithm that improves state-of-the-art performance for object classification of cars. As classification rates near 100\% for the N-CARS, the lack of large labeled datasets will limit advancement in this area. Multiple simulators now exist for generating synthetic data \cite{rebecq2018esim} \cite{mueggler2017event}, which have been used successfully in several papers for testing. Although these simulators may be useful in the short term, real-world data is always preferred as noise, calibration, and manufacturing defects are challenging to reliably simulate.

Two limitations of IETS should be addressed with future work. First, IETS relies on the fact that edges triggering events rarely generate large, overlapping time-surfaces within 100 milliseconds. This may not be true for all scenarios. For example, a spinning fan, pulsing light, or very fast moving object would generate overlapping surfaces and likely limit the utility of IETS in these cases. The IETS algorithm currently averages overlapping surfaces, but this is not optimal as these unique signatures are undetectable to a standard camera. Second, after the time-surfaces are generated from IEs, no effort is made to recover data originally filtered as noise. A two-stage filter design will help recover events and allow for a broader application of the algorithm.

%
%
%
\bibliographystyle{splncs04}
\bibliography{egbib}
\end{document}